\pdfoutput=1

\documentclass[11pt]{article}

\usepackage{ACL2023}

\usepackage{times}
\usepackage{latexsym}
\usepackage{graphicx}
\usepackage{booktabs}
\usepackage{multirow}
\usepackage{tikz}
\usetikzlibrary{tikzmark}
\usepackage{color, colortbl}
 \definecolor{LightCyan}{rgb}{0.8,1,1}

\usepackage[utf8]{inputenc}
\usepackage{microtype}

\usepackage{inconsolata}
\usepackage{xspace}
\usepackage{diagbox}
\newcommand{\org}{{\sc {\tt ORG}}\xspace}
\newcommand{\loc}{{\sc {\tt LOC}}\xspace}
\newcommand{\comp}{{\sc {\tt COMP}}\xspace}
\newcommand{\perderiva}{{\sc {\tt PERderivA}}\xspace}
\newcommand{\perderiv}{{\sc {\tt PERderiv}}\xspace}
\newcommand{\per}{{\sc {\tt PER}}\xspace}
\newcommand{\oth}{{\sc {\tt OTH}}\xspace}
\usepackage{subcaption}

\usepackage{xparse}
\usepackage{tikz}
\usetikzlibrary{arrows.meta}
\usetikzlibrary{bending}
\usetikzlibrary{calc}
\usetikzlibrary{matrix}

\NewDocumentCommand\treeword{m m m}{
	\node[align=center, inner sep=0] (#1) {\strut #1\\[-1.3mm]\scriptsize\textcolor{violet}{\textsc{#2}}\\[-1mm]\footnotesize\strut\textit{#3}};
	\coordinate (#1-in) at (#1.north);
	\coordinate (#1-out-right) at ($(#1.north)+(-1mm,0)$);
	\coordinate (#1-out-left) at ($(#1.north)+(1mm,0)$)
}

\NewDocumentCommand\treeroot{m m}{
	\draw[-{Latex}] ($(#1-in)+(0,#2mm)$) to node[pos=0, right, anchor=north west] {\footnotesize\textcolor{violet}{root}} (#1-in);
}

\NewDocumentCommand\treerel{m m m m m}{
	\draw[-{Latex[flex]}] (#1-out-#3) to[bend #3=90, distance=#5mm] node[midway, above, inner sep=0.4mm] {\footnotesize\textcolor{violet}{#4}} (#2-in);
}

\expandafter\def\expandafter\UrlBreaks\expandafter{\UrlBreaks
  \do\a\do\b\do\c\do\d\do\e\do\f\do\g\do\h\do\i\do\j%
  \do\k\do\l\do\m\do\n\do\o\do\p\do\q\do\r\do\s\do\t%
  \do\u\do\v\do\w\do\x\do\y\do\z\do\A\do\B\do\C\do\D%
  \do\E\do\F\do\G\do\H\do\I\do\J\do\K\do\L\do\M\do\N%
  \do\O\do\P\do\Q\do\R\do\S\do\T\do\U\do\V\do\W\do\X%
  \do\Y\do\Z}

\usepackage{soulutf8}
\usepackage[normalem]{ulem}
\usepackage{xcolor}

\usepackage{enumitem}
\setlist{nolistsep}
\usepackage{microtype}

\iftrue 
    \usepackage[final]{proofing}
  \renewcommand\hl[1]{{#1}}  
   {\draftnote{\red{#2}}}
   \newcommand\redHL[1]{}
  \newcommand\todo[1]{}

  \newcommand{\Djame}[1]{}

\else  

\usepackage[draft]{proofing}

\newcommand{\Djame}[1]{
\textbf{\textcolor{red}{\hl{Djame: #1}}}
}

\newcommand\red[1]{{{\textcolor{red}{\bf #1}}}}

\let\oldred\red
\renewcommand\red[1]{{ \oldred{{#1}}}}

 \newcommand\redHL[1]{\red{\hl{#1}}}
\let\olddraftnote\draftnote
\renewcommand\draftnote[1]{\olddraftnote{\red{#1}}}

\fi

\title{Enriching the NArabizi Treebank: A Multifaceted Approach to Supporting an Under-Resourced Language}

\author{Arij Riabi \quad Menel Mahamdi \quad Djamé Seddah \\
     Inria, Paris\\
     \{firstname,lastname\}@inria.fr}

\usepackage{listings}

\begin{document}






\maketitle
\begin{abstract}

  In this paper we address the scarcity of annotated data for
  NArabizi, a Romanized form of North African Arabic used mostly on
  social media, which poses challenges for Natural Language Processing
  (NLP). We introduce an enriched version of NArabizi Treebank
  \cite{seddah-etal-2020-building} with three main contributions: the
  addition of two novel annotation layers (named entity recognition
  and offensive language detection) and a re-annotation of the
  tokenization, morpho-syntactic and syntactic layers that ensure
  annotation consistency. Our experimental results, using different
  tokenization schemes, showcase the value of our contributions and
  highlight the impact of working with non-gold tokenization for NER
  and dependency parsing.  To facilitate future research, we make
  these annotations publicly available. Our enhanced NArabizi Treebank
  paves the way for creating sophisticated language models and NLP
  tools for this under-represented language.

\end{abstract}

\section{Introduction}
\iffalse Even though the world abounds with many rich and diverse
dialects, each possessing distinct features and characteristics, many
of these dialects still lack the resources and support required for
their speakers to access contemporary technologies in their language
\cite{joshi-etal-2020-state}.  Therefore, efforts to create annotated
corpora, develop language models, or establish dictionaries and
grammars for low-resource dialects are critical for preserving and
promoting these dynamic languages, which capture the community's
unique cultures, histories, and experiences. For example, the
Masakhane community aims to strengthen NLP research for African
languages with important initiatives such as MasakhaNER
\cite{adelani-etal-2021-masakhaner}. By working on resources for
low-resource dialects, we contribute to their preservation and
promotion.  
\else 

Despite the abundance of rich and diverse dialects
worldwide, each possessing distinctive features and characteristics,
many of these dialects still lack the necessary resources and support
to enable their speakers to access modern technologies in their own
language \cite{joshi-etal-2020-state}. Therefore, it is imperative
to undertake endeavors aimed at creating annotated corpora, developing
language models, and establishing dictionaries and grammars for
low-resource dialects. These efforts are crucial for the preservation
and advancement of these dynamic languages, which encapsulate unique
cultures, histories, and experiences within their respective
communities. 

One notable example of such an effort is the Masakhane community, which is
dedicated to enhancing natural language processing (NLP) research for
African languages through significant initiatives such as MasakhaNER
\cite{adelani-etal-2021-masakhaner}. Similar efforts are ongoing for
Indonesian languages \cite{cahyawijaya2022nusacrowd}.

\iffalse 

Another long standing, orthogonal in a way, established initiative
that initially aimed at offering a common set of syntactic guidelines
for a small set of languages, the Universal Dependencies project
\cite{nivre-etal-2020-universal}, turned out to be the recipient of
many low-resources languages treebank initiatives that followed and
often extended these initial guidelines to cope with their respective
idiosyncrasies.

\else 
In addition, a long-standing and somewhat unrelated initiative
known as the Universal Dependencies project
\cite{nivre-etal-2020-universal} originally aimed to provide a
standardized set of syntactic guidelines for a limited number of
languages turned out to become the recipient of numerous
treebank initiatives for low-resource languages. These initiatives not
only adopted the initial guidelines but also expanded upon them to
accommodate the unique idiosyncrasies of each language.  \fi

In this work, we aim to enhance a pre-existing multi-view
treebank  devoted to a very low-resource language, namely the North-African Arabic dialect written in
Latin script, collected from Algerian sources and denoted as the
Narabizi treebank, the first
 available for this dialect, where Arabizi refers to both the practice of
 writing Arabic using the Latin alphabet and \textit{N} for the North
 African dialect  \cite{seddah-etal-2020-building}. 
Made of noisy user-generated content that exhibits a high level of
language variability, its annotations faced many challenges as
described by the authors and contained remaining errors \cite{touileb-barnes-2021-interplay}. 

Our work builds on previous efforts to annotate and standardize
treebank annotations for low-resource languages to enhance the quality
and consistency of linguistic resources
\cite{schluter2007preparing,sade-etal-2018-hebrew,turk-etal-2019-turkish,zariquiey-etal-2022-building}.

 Following previous research, we consider the impact of refining
 annotation schemes on downstream tasks.
 \citet{mille-etal-2012-granularity} examine how much a treebank's
 performance relies on its annotation scheme and whether employing a
 more linguistically rich scheme would decrease performance. Their
 findings indicate that using a fine-grained annotation for training a
 parser does not necessarily improve performance when parsing with a
 coarse-grained tagset. This observation is relevant to our study as
 we expect refining the treebank could enhance the parsing
 performance even though the inherent variability of this language,
 which, tied to its  small size treebank, could bring a negative impact on such
 enhancements.  

 On the other hand, the experiments conducted by
 \citet{schluter2007preparing} demonstrate that using a cleaner and
 more coherent treebank yields superior results compared to a treebank
 with a training set five times larger. This observation highlights
 the significance of high-quality dataset annotations, particularly
 for smaller datasets. This understanding primarily drives the goal of
 improving the NArabizi treebank's annotations.

 In this context, we propose a heavily revised version of NArabizi
 treebank \cite{seddah-etal-2020-building} that includes two novel
 annotation layers for Named Entity Recognition (NER) and offensive
 language detection. One of the goals of this work is also to study the
 impact of non-gold tokenization on NER, a scenario almost never investigated
by the community \cite{bareket-tsarfaty-2021-neural}.
Our primary contributions are as follows:

\begin{itemize}
    \item Using error mining tools, we release a new corrected version
      of the treebank, which leads to improved downstream task performance.
    \item We show that corrections made to a small size treebank of a
      highly variable language favorably impacts the performance of NLP models trained on it.
    \item We augment the treebank by adding NER annotations and offensive language detection, expanding its applicability in various NLP tasks.
    \item We homogenize tokenization across the dataset, analyze the
      impact of proper tokenization on UD tasks and NER and conduct a
      realistic evaluation on predicted tokenization, including 
      NER evaluation.
\end{itemize}

The enhanced version of the Narabizi Treebank is freely available.\footnote{\url{https://gitlab.inria.fr/ariabi/release-narabizi-treebank}} 

\section{Related work}
\paragraph{NArabizi}
The Arabic language exhibits diglossia, where Modern Standard Arabic (MSA) is employed in formal contexts, while dialectal forms are used informally \cite{habash10intro}. Dialectal forms, which display significant variability across regions and predominantly exist in spoken form, lack standardized spelling when written. Many Arabic speakers employ the Latin script for transcribing their dialects online, using digits and symbols for phonemes not easily mapped to Latin letters \cite{seddah-etal-2020-building}. This written form, known as Arabizi and its North African variant, NArabizi, often showcases code-switching with French and Amazigh \cite{amazouz2017addressing}. Textual resources for Arabizi primarily consist of noisy, user-generated content \cite{foster-2010-cba,seddah-etal-2012-french,eisenstein2013bad}, complicating the creation of supervised models or collection of extensive pre-training datasets. The original NArabizi treebank  \cite{seddah-etal-2020-building}, contains about 1500 sentences. The sentences are randomly sampled from the romanized Algerian dialectal Arabic corpus of \citet{cotterell2014algerian} and from a small corpus of lyrics from Algerian dialectal Arabic songs popular among the younger generation. This treebank is manually annotated with morpho-syntactic information (parts-of-speech and morphological features), together with glosses and code-switching labels at the word level, as well as sentence-level translations to French. Moreover, this treebank also contains 36\% of French tokens. \draftadd{Since its creation, this treebank spawned two derived versions that first added a transliteration to the Arabic script at the word level and sentiment and topic annotation at the sentence level \cite{touileb-barnes-2021-interplay}. In parallel to our own corrections and annotation work\footnote{Released on November 26th, 2022, the same day as the publication of \cite{touileb2022nerdz}.}, \citet{touileb2022nerdz}} extended this work to include a named-entity annotation layer.

\paragraph{Treebanking for User-generated Content}

Treebanks and annotated corpora have greatly impacted NLP tools, applications, and research in general. Despite the challenges of constructing large and structurally consistent corpora, which requires considerable effort and time, many in the field considered this pursuit valuable and necessary \cite{de-marneffe-etal-2021-universal}. However, constructing treebanks for user-generated content is more challenging due to the extensive variation in language usage and style, the prevalence of non-standard spellings and grammar, and the necessity for domain-specific annotations \cite{sanguinetti2022treebanking}. Interest in treebanking user-generated content, such as social media posts and online forum discussions, has risen, and numerous efforts have been undertaken to create treebanks for user-generated content \cite{foster-etal-2011-news,seddah-etal-2012-french,sanguinetti-etal-2018-postwita,rehbein-etal-2019-tweede,sanguinetti-etal-2020-treebanking}.
\paragraph{NER for Dialects and User-generated Content}

NER is an information extraction task that identifies and categorizes entities at the token level. It is an extensively investigated NLP task with numerous datasets and models for various languages. However, datasets for low-resource languages are rare, and NER datasets for social media platforms such as Twitter predominantly exist for English \cite{ritter-etal-2011-named,derczynski-etal-2016-broad,derczynski-etal-2017-results}.

A prominent NER dataset for {\em lower-than-English} resource languages is the CoNLL 2002 Shared Task dataset \cite{tjong-kim-sang-2002-introduction}, which provides NER annotations for four languages: Dutch, Spanish, Chinese, and Czech. Additionally, the WikiAnn dataset \cite{pan-etal-2017-cross} includes NER annotations for several low-resource languages. Nevertheless, it is derived from Wikipedia content which is not well-suited for NER tasks involving user-generated content. As mentioned above, \citet{touileb2022nerdz} added a NER annotation for the first version of the NArabizi treebank. However, they did not address the tokenization issues inherent in the dataset and used a different annotation scheme. The following sections delve deeper into the tokenization challenges and the differences between the two datasets. 

\section{Extending a Low-resource Language treebank}
In this section, we outline our methodology for expanding and enhancing the NArabizi treebank. We start by re-annotating tokenization, morpho-syntactic, and syntactic layers to ensure consistency, followed by detailing the annotation guidelines and procedures for NER and Offensive Language detection. We refer to the initial treebank introduced by \citet{seddah-etal-2020-building} as NArabiziV1 and our extended version as NArabiziV2.
\subsection{Maintaining Consistency in Treebank Annotations}
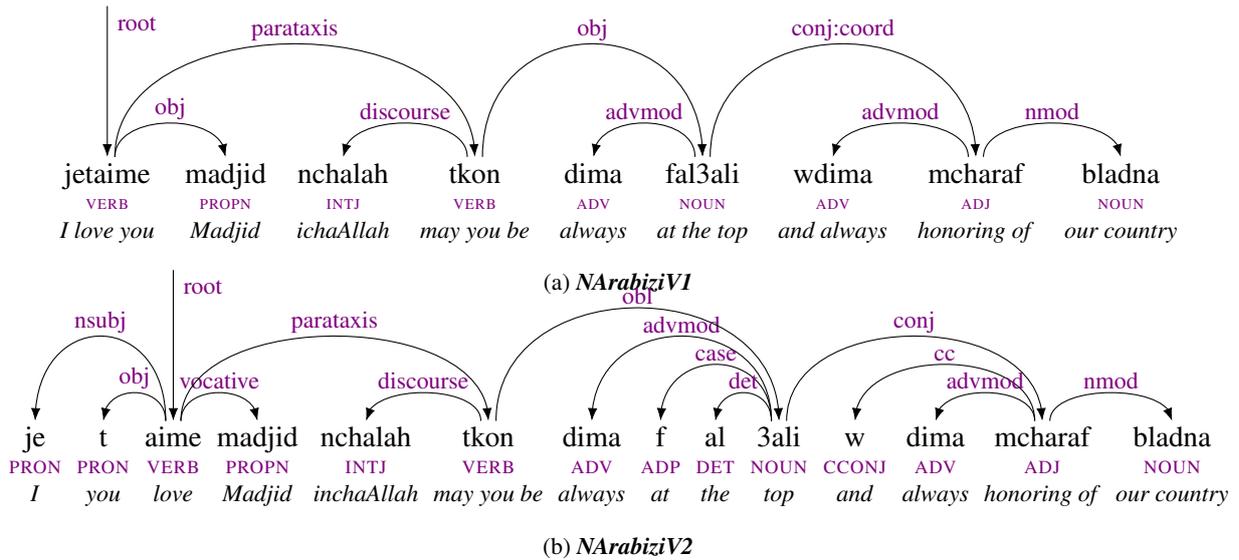
\begin{figure*}
\centering
\begin{tikzpicture}
\begin{scope}
	\matrix[column sep=4mm] (sentence) {
		\treeword{jetaime}{verb}{I love you}; &
		\treeword{madjid}{propn}{Madjid}; &
		\treeword{nchalah}{intj}{ichaAllah}; &
		\treeword{tkon}{verb}{may you be}; &
		\treeword{dima}{adv}{always}; &
		\treeword{fal3ali}{noun}{at the top}; &
		\treeword{wdima}{adv}{and always}; &
		\treeword{mcharaf}{adj}{honoring of}; &
		\treeword{bladna}{noun}{our country}; \\
	};

	\treeroot{jetaime}{20};
	\treerel{jetaime}{madjid}{left}{obj}{6};
	\treerel{jetaime}{tkon}{left}{parataxis}{20};
	\treerel{fal3ali}{dima}{right}{advmod}{6};
	\treerel{tkon}{fal3ali}{left}{obj}{20};
	\treerel{tkon}{nchalah}{right}{discourse}{6};
	\treerel{fal3ali}{mcharaf}{left}{conj:coord}{20};
	\treerel{mcharaf}{wdima}{right}{advmod}{6};
	\treerel{mcharaf}{bladna}{left}{nmod}{6};
\node [below of=sentence] {\parbox{0.3\linewidth}{\subcaption{\textbf{\textit{NArabiziV1}}}}};
\end{scope} 
\begin{scope}[yshift=-35mm]
	\matrix[column sep=2mm] (sentence) {
		\treeword{je}{PRON}{I}; &
		\treeword{t}{PRON}{you}; &
		\treeword{aime}{VERB}{love}; &
		\treeword{madjid}{PROPN}{Madjid}; &
		\treeword{nchalah}{INTJ}{inchaAllah}; &
		\treeword{tkon}{VERB}{may you be}; &
		\treeword{dima}{ADV}{always}; &
		\treeword{f}{ADP}{at}; &
		\treeword{al}{DET}{the}; &
		\treeword{3ali}{NOUN}{top}; &
		\treeword{w}{CCONJ}{and}; &
		\treeword{dima }{ADV}{always}; &
		\treeword{mcharaf}{ADJ}{honoring of}; &
		\treeword{bladna}{NOUN}{our country}; \\
	};

	\treeroot{aime}{20};
	\treerel{aime}{je}{right}{nsubj}{15};
	\treerel{aime}{t}{right}{obj}{5};
	\treerel{aime}{madjid}{left}{vocative}{5};
	\treerel{tkon}{nchalah}{right}{discourse}{5};
	\treerel{aime}{tkon}{left}{parataxis}{15};
	\treerel{3ali}{dima}{right}{advmod}{15};
	\treerel{3ali}{f}{right}{case}{10};
	\treerel{3ali}{al}{right}{det}{5};
	\treerel{tkon}{3ali}{left}{obl}{20};
	\treerel{mcharaf}{w}{right}{cc}{10};
	\treerel{mcharaf}{dima }{right}{advmod}{5};
	\treerel{3ali}{mcharaf}{left}{conj}{15};
	\treerel{mcharaf}{bladna}{left}{nmod}{5};
\end{scope}
\node [below of=sentence] {\parbox{0.3\linewidth}{\subcaption{\textbf{\textit{NArabiziV2}}}}};
\end{tikzpicture}
\caption{\label{tree}
Illustration of an example from the NAarabizi treebank before and after the modifications.
}
\end{figure*}
We start with an extended clean-up of the NArabiziV1 formatting, which involves reinstating missing part-of-speech tags and rectifying Conllu formatting discrepancies. Then, we embark on general error mining in the lexical and syntactical annotation and correction phase. We implement this stage using semi-automated methods. We do not change the UD tagsets used in the original treebank.

\paragraph{Error Mining} We use the UD  validator Vr2.11 \footnote{\url{https://github.com/UniversalDependencies/tools/releases/tag/r2.11}}, a tool designed to assess the annotation of treebanks in UD and ensure compliance with the UD specifications. The validator is specifically employed to detect common errors, such as invalid dependency relations, incorrect part-of-speech tags, and inconsistent usage of features like tense and aspect. By leveraging the UD validator, we guarantee that our dataset is syntactically consistent and conforms to the standards established by the UD project. These changes encompass correcting cycle and projectivity issues and removing duplicates.

We also use Errator \cite{wisniewski2018errator}, a data mining tool, to pinpoint inconsistencies in our dataset.
It implements the annotation principle presented by \citet{boyd2008detecting}, which suggests that if two identical word sequences have different annotations, one is likely erroneous.

We remove the duplicated sentences when the text field is an exact match and fix duplicated sentence identification for different sentences. We also fixed some problems with the original text, such as Arabic characters encoding and sentence boundaries.  

\paragraph{Tokenization} We address tokenization concerns to uphold consistency in the NArabizi Treebank annotations. Furthermore, we introduce targeted adjustments to resolve issues related to segmenting specific word classes, including conjunctions, interjections (e.g., ``ya''), determiners, and prepositions, especially when adjacent to noun phrases. For example, we segment determiners at the initial vowel (``a'' or ``e''), as demonstrated in the examples ``e ssalam'' (``the peace'') and ``e dounoub'' (``the sins''). The lemma field for these terms is aligned with the French translation for the splitting (e.g., ``e ssalam'' $\Rightarrow$ ``la paix'' (``the peace'')). For prepositions, we perform splitting at the first letter followed by ``i'' when possible, as seen in ``brabi'' $\Rightarrow$ ``b rabi'' (``with my god''). We also establish rules for segmenting determiners and proper nouns. When possible, we separate prepositions at the initial letter and ``i'' and instituted guidelines for segmenting determiners and proper nouns. We implement these alterations for splitting using the Grew graph rewriting tool for NLP \cite{guillaume2021graph} to improve the consistency and quality of the treebank annotations. Additionally, we fix all the problems mentioned by \citet{touileb-barnes-2021-interplay} regarding the incoherence of the tokenization, wrong translations, and incoherent annotations.

\paragraph{Translation} The translation quality is also enhanced; previously, translations were not consistently carried out by Algerian speakers, resulting in local expressions and phrases being frequently misinterpreted, either in a literal manner or, at times, entirely inaccurately. This had implications for lexical and syntactical annotation. For instance, the term ``skara'' was initially annotated as ``on purpose" but was later revised to ``taunting''. Recognizing that ``skara fi'' represents a local expression facilitates annotation and promotes corpus harmonization.

\paragraph{Example} In Figure \ref{tree}, we illustrate a parse tree before and after applying several corrections. Tokenization errors in French were rectified (``jetaime'' $\Rightarrow$ ``je t aime''), and Arabic prepositions, articles, and conjunctions were separated from the nouns or adverbs they were attached to (``fal3ali'' $\Rightarrow$ ``f al 3ali'', ``wdima'' $\Rightarrow$ ``w dima''). We also correct some dependency relations: the previous ``obj'' relation between the verb ``aimer'' and the proper noun ``madjid'' was altered to ``vocative''. 


\paragraph{Interesting Properties} The corpus displays several interesting linguistic features, including \textit{parataxis}, \textit{goeswith}, and dislocated structures, characteristic of oral productions and user-generated content. A deeper examination of the root/parataxis ratio and the average parataxis per tree in the corpus, which contains 2066 parataxis for 1287 sentences, shows \draftreplace{that the corpus exhibits a high level of syntactic complexity}{that the corpus exhibits a high level of juxtaposed clauses resulting from the absence of punctuation. Given the initial data sources (web forums), it is likely that these end of sentences markers were initially present as carriage returns}.

As pointed out by \citet{seddah-etal-2020-building} the corpus also exhibits a high level of spelling variation, reflecting the speakers' diversity in terms of geography and accents.\draftnote{changé -ds} Furthermore, analyzing the number of sentences without a verb and the average number of verbs per sentence shows that NArabizi speakers tend to favor nominalization, as seen in the abundance of ellipses (e.g., ``rabbi m3ak'' which translates in English to ``God bless you'').

\subsection{Annotation Methodology for NER and Offensive Language Detection}
\paragraph{Named Entity Recognition}
Our NER annotation guidelines are based on the revised tokenization of the NArabizi treebank, which ensures consistency between token-level annotations, an essential aspect of multi-task learning. We use the Inception tool \cite{klie2018inception} for our manual annotation by two native speakers, adhering to the IOB2 Scheme \cite{tjong-kim-sang-veenstra-1999-representing}. Each word is labeled with a tag indicating whether it is at the beginning, inside, or outside of a named entity. In case of disagreement between annotators, the multiple annotations were subsequently discussed until agreement was reached, and one annotation was selected to be retained.  
We extend the FTB NER \cite{sagot2012annotation} French treebank annotations. Our annotation contains the following NE types: \per for real or fictional persons, \org for organizations, \loc for locations, \comp for companies, and \oth for brands, events, and products. 

In cases of ambiguity between products and companies, we adhere to the decision made in the FTB dataset. For person names, we exclude grammatical or contextual words from the mention. We annotate football teams as organizations, and we annotate mentions of "Allah" or "Rabi" as \perderiva. The \perderiv annotation is applied to groups of individuals who originate from or share the same location. Country names are consistently labeled as locations, irrespective of the context. TV channels and ambiguous brand names are annotated as companies, while religious groups are not designated entities. The names of football stadiums are classified under \oth, whereas journal names are identified as organizations.

Table \ref{tab:stats-corpus-ner} presents the distribution of entities, with a similar distribution observed across both the development and test splits. The most frequent entity type is \perderiva, while the least frequent is \comp.

\begin{table}[!h]
    \centering
    \footnotesize
    \begin{tabular}{lrrrr}
    \toprule
    Type &   train & dev & test &Total \\
    \midrule
    \per  & 371 & 61 & 47 &479\\
    \loc  &  358 & 58 & 50 &466\\
    \org  & 200 &23 &28 &251\\
    \comp &6  &5 &3 &14\\
    \oth  & 44 & 6&7 &57\\
    \perderiv  & 96& 14&13 &123\\
    \perderiva  &386 & 57 &66 & 509\\
    \midrule
    Total & 1461 & 224 & 214 &1899\\
    \bottomrule
    \end{tabular}
    \caption{Named entity type distribution across train, dev,
and test splits.}
    \label{tab:stats-corpus-ner}
\end{table}

\begin{table}[!h]
    \centering
    \footnotesize
    \begin{tabular}{lrrr}
    \toprule
    Type &   train & dev & test  \\
    \midrule
    nb sentences  & 1003 & 139 & 145\\
    nb tokens  &  15522  &  2124 & 2118 \\
    nb unique tokens  &  6652 & 1284  & 1327 \\
    \bottomrule
    \end{tabular}
    \caption{Statistics of the deduplicated corpus across train, dev,
and test splits. The train-dev intersection contains 549 tokens, the train-test intersection contains 551 tokens, and the dev-test intersection contains 266 tokens.}
    \label{tab:stats-corpus}
\end{table}
Table \ref{tab:stats-corpus} displays the number of unique words which can provide information about the language used in the corpus. The fact that the count of unique tokens constitutes nearly half of the total tokens suggests that the language used in the corpus is complex and diverse, with a wide range of vocabulary and expressions. This can make it more challenging for NER algorithms to accurately identify and classify named entities in the corpus.

\citet{touileb2022nerdz} recently introduced NERDz, a version of the NArabizi treebank annotated for NER. As our dataset's annotation labels differ from theirs, we establish a mapping between the two annotation schemes to enable comparisons (cf. see Table \ref{tab:comp-nerdz} in the appendix \ref{sec:appendix}). Our schemes also differ in named entities' scope, as we split contracted forms, ours only cover the nominal phrase parts. Regarding nouns, such as ``\textit{bled}'', which means {\em country}, some are annotated as entity GPE in NERDz, which is not the case in our dataset. Also, the names of stadiums are annotated as \loc in NERDz while they are considered \oth in our dataset. Similarly, for ``\textit{equipe nationale}'', which means {\em national team} is annotated \org in NERDz, while we do not consider it as an entity, following the FTB NER's guidelines. Added to annotator divergences, this may explain the differences in the count of the entities.

\paragraph{Offensive Language Classification}
The annotation process for offensive language classification was conducted manually by three annotators with diverse backgrounds. The annotators consisted of two females and one male, each bringing unique expertise to the task. One female annotator is a Ph.D. student in NLP, the other is a Ph.D. student in political sciences, and the male annotator is an engineer with in-depth knowledge of North African football, a prominent topic in the dataset.

The annotators were asked to annotate every sentence as offensive (OFF) or non-offensive (NOT-OFF). Offensive posts included any form of unacceptable language, targeted offense (veiled or direct), insults, threats, profane language, and swear words.
To maintain objectivity and minimize potential bias, the annotators were not granted access to the other annotators' work and were not allowed to discuss their annotations with one another. This approach ensured the independence of their judgments, allowing for a more reliable evaluation of the offensive language classification process.
For the offensive annotation, the two female annotators did not usually agree with the male annotator as they have different backgrounds and hence different opinions about football-related sentences.
The final label is determined through a majority voting process. Additionally, we calculate the average pair-wise Cohen's $\kappa$ \cite{cohenkappa} to highlight how hard this task was. The average $\kappa$ value is 0.54, indicating a moderate agreement between  annotators, common in sentence level annotation for annotators with different backgrounds and topic familiarity \cite{bobicev-sokolova-2017-inter}.
This disagreement likely stems from the interpretation of terms that can be considered offensive or non-offensive depending on either the dialect or context.

Table \ref{tab:dataset_offensive} presents the distribution of non-offensive and offensive language instances. The dataset features an imbalance between non-offensive and offensive classes, with non-offensive samples being considerably more frequent in each split.
\begin{table}[ht]
\footnotesize
\centering
\begin{tabular}{lcc}
\toprule
Split & Non-Offensive & Offensive \\
\midrule
Train   & 804           & 199       \\
Dev     & 86            & 53        \\
Test    & 118           & 27        \\
\bottomrule
\end{tabular}
\caption{Offensive language detection distributions across train, dev, and test splits.}
\label{tab:dataset_offensive}
\end{table}

\section{Dataset Evaluation}
We evaluate the NarabiziV2 dataset on  UD parsing tasks and NER using standard transfer learning architectures on which we vary the pre-trained language model and the tokenization scenario. 

\paragraph{New NArabizi CharacterBert Model}
Following \citet{riabi-etal-2021-character}, we train a CharacterBERT \cite{el-boukkouri-etal-2020-CharacterBERT} model, a character-based BERT variant, on a NArabizi new filtered corpus. The authors demonstrate that CharacterBERT achieves significant results when dealing with noisy data while being extremely data efficient. \draftremove{They also show when sufficient data is unavailable for training a large language model, a CharacterBERT model trained on limited corpora can still yield competitive performances.}
 
We improve the initial pre-training dataset used by \citet{riabi-etal-2021-character} by more stringently filtering non-NArabizi examples from the 99k instances provided by \citet{seddah-etal-2020-building}, as well as incorporating new samples from the CTAB corpus \cite{amina_amara_2021_4780941} and 12k comments extracted from various Facebook and forum posts, mostly in the Tunisian dialect taken from different datasets listed by \citet{younes2020language}. This results in a 111k sentence corpus. To exclude non-NArabizi content, we first use a language detection tool \cite{nakatani2010langdetect} with a 0.9 confidence threshold to eliminate text in French, English, Hindi, Indonesian, and Russian, which are commonly found in mixed Arabizi data. Following the filtering process, a bootstrap sampling method is adopted to randomly select a subset of the remaining text for manual annotation. This annotated text is then used to train an SVM classifier for NArabizi detection. The final dataset, containing 91k annotated text instances after deduplication, focuses on North African Arabizi text. We make this corpus publicly available.

\paragraph{Sub-word Models}
We also evaluate the performance of subword-based language models, monolingual and multilingual. For the multilingual subword-based language model, we use mBERT, the multilingual version of BERT \cite{devlin2018bert}. It is trained on data from Wikipedia in 104 different languages, including French and Arabic. \citet{muller2020can} demonstrated that such a model could be transferred to NArabizi to some degree. Finally, our monolingual model is DziriBERT \cite{abdaoui2021DziriBERT}, a monolingual BERT model trained on 1.2M tweets from major and highly-populated Algerian cities scrapped using a set of popular keywords in the Algerian spoken dialect in both Arabic and Latin scripts.
\section{Results}
 \begin{table*} [h!]
\centering
\footnotesize
\begin{tabular}{@{}llccc|ccc@{}}
\toprule
\multirow{2}{*} {Model}& \multirow{2}{*}{\backslashbox{\scriptsize{Train}\kern-1.5em}{\kern-1.5em \scriptsize{Test}}} & \multicolumn{3}{c}{\textit{NArabiziV1}}                                               & \multicolumn{3}{c@{}}{\textit{NArabiziV2}}                                            \\ 
\cline{3-8}
&  & \rule{0pt}{2.5ex}{UPOS} & {UAS} & {LAS} & {UPOS} & {UAS} & {LAS}  \\ 
\hline
     \rule{0pt}{2.5ex} mBERT & \multirow{3}{*}{\rotatebox[origin=c]{90}{\textit{{\centering \scriptsize{NArabiziV1}}}}}       & \tikzmark{k}77.42 \textsuperscript{$\pm$ 1.52} & 68.91 \textsuperscript{$\pm$ 0.65} & 56.19 \textsuperscript{$\pm$ 0.86}      & 74.59 \textsuperscript{$\pm$ 1.42} & 66.01 \textsuperscript{$\pm$ 0.47} & 53.19 \textsuperscript{$\pm$ 0.87}  \\
                        \rule{0pt}{2.5ex} DziriBERT  & & 83.57 \textsuperscript{$\pm$ 0.92} & 73.97 \textsuperscript{$\pm$ 0.72} & 62.04 \textsuperscript{$\pm$ 0.54}     &\cellcolor{LightCyan}{ 80.19 \textsuperscript{$\pm$ 0.82}} & \cellcolor{LightCyan}{70.28 \textsuperscript{$\pm$ 0.83} }& \cellcolor{LightCyan}{58.63 \textsuperscript{$\pm$ 0.78}   }    \\
                        \rule{0pt}{2.5ex} CharacterBERT  &  & 76.19 \textsuperscript{$\pm$ 2.48} & 68.78 \textsuperscript{$\pm$ 0.36} & 55.14 \textsuperscript{$\pm$ 0.38}  \tikzmark{8}  & 73.01 \textsuperscript{$\pm$ 2.05} & 66.10 \textsuperscript{$\pm$ 0.48} & 52.41 \textsuperscript{$\pm$ 0.50} \\
                        \midrule
    \rule{0pt}{2.5ex} mBERT & \multirow{3}{*}{\rotatebox[origin=c]{90}{\textit{{\centering \scriptsize{NArabiziV2}}}}}       & 74.48 \textsuperscript{$\pm$ 0.95} & 66.03 \textsuperscript{$\pm$ 0.35} & 52.82 \textsuperscript{$\pm$ 0.66}   & \tikzmark{J} 79.65 \textsuperscript{$\pm$ 0.90} & 70.56 \textsuperscript{$\pm$ 0.32} & 58.08 \textsuperscript{$\pm$ 0.76}  \\
                       \rule{0pt}{2.5ex}  DziriBERT &   & 78.75 \textsuperscript{$\pm$ 1.29} & 70.51 \textsuperscript{$\pm$ 0.43} & 57.51 \textsuperscript{$\pm$ 0.67}    & \cellcolor{LightCyan}{83.10 \textsuperscript{$\pm$ 1.60} }& \cellcolor{LightCyan}{74.26 \textsuperscript{$\pm$ 0.27} }& \cellcolor{LightCyan}{62.66 \textsuperscript{$\pm$ 0.52}}\\                    
                       \rule{0pt}{2.5ex} CharacterBERT  &  & 72.24 \textsuperscript{$\pm$ 2.62} & 65.74 \textsuperscript{$\pm$ 0.24} & 51.86 \textsuperscript{$\pm$ 0.51}   & 76.34 \textsuperscript{$\pm$ 2.68} & 69.84 \textsuperscript{$\pm$ 0.27} & 56.27 \textsuperscript{$\pm$ 0.54} \tikzmark{7}    \\
\bottomrule
\end{tabular}
\begin{tikzpicture}[overlay,remember picture]
  \draw[red] ([shift={(-1ex,2ex)}]pic cs:J) rectangle ([shift={(1ex,-0.5ex)}]pic cs:7);
   \draw[blue] ([shift={(-1ex,2ex)}]pic cs:k) rectangle ([shift={(1ex,-0.5ex)}]pic cs:8);
\end{tikzpicture}
\caption{Results for UD on test set, DEV set is used for validation (with gold tokenization) (We report average of F1 scores over 5 seeds with the standard deviation)}
\label{table:goldtokenparsing}
\end{table*}
\subsection{New Results for UD}
For our updated version of the treebank, we present results for models trained and tested on NArabiziV2, as shown in Table \ref{table:goldtokenparsing} and highlighted by a red box. These results represent the new state-of-the-art performance for the treebank, and we report findings for three previously used models. The DziriBERT model exhibits the best performance; however, CharacterBERT delivers competitive results while being trained on a mere 7.5\% of the data used for training DziriBERT. This observation is consistent with the conclusions drawn by \citet{riabi-etal-2021-character}.

In order to assess the influence of the implemented corrections, we use NArabiziV1 and eliminate duplicate sentences
\footnote{To use the prior version with an equivalent number of sentences, format errors must be rectified (earlier experiments with these sentences excluded them).}. For this comparison, we focused on the DziriBERT model's performance when trained on either NArabiziV1 or NArabiziV2 and tested on NArabiziV2, as denoted by the blue highlights in Table \ref{table:goldtokenparsing}. Training on NArabiziV2 enhances the average scores for UPOS, UAS, and LAS by 3.5 points, illustrating the favorable outcomes of the refinements introduced in the NArabiziV2 dataset. This observation is further substantiated by examining the performance of CharacterBERT and mBERT, reinforcing the validity of the noted improvements.

A comparative analysis of the results for models trained and tested on NArabiziV1, denoted by the blue box, and those for models trained and tested on NArabiziV2, denoted by the red box, reveals that NArabiziV2 generally yields superior evaluation scores. This observation underlines the impact of the treebank's consistency on the overall performance of the models. When we test on NArabiziV1, the model trained on NArabiziV1 gets better results than the model trained on NArabiziV2. The modifications in tokenization can explain this drop in performance.

\subsection{Results for NER and Offensive Language Detection}
\begin{table*}[h!]
\centering
    \footnotesize
\begin{tabular}{lccccccccc}
\toprule
Model & \loc & \org & \per & \oth & \perderiv & \perderiva & macro avg\\
\midrule
mBERT & 82.93 \textsuperscript{$\pm$ 4.02} & 66.17 \textsuperscript{$\pm$ 6.61} & 61.84 \textsuperscript{$\pm$ 3.56} & 25.56 \textsuperscript{$\pm$ 14.64} & 57.98 \textsuperscript{$\pm$ 11.30} & 95.62 \textsuperscript{$\pm$ 1.24}  & 65.02 \textsuperscript{$\pm$ 1.24}\\
DziriBERT & 85.84 \textsuperscript{$\pm$ 3.43} & \textbf{73.67} \textsuperscript{$\pm$ 4.03} & \textbf{73.42} \textsuperscript{$\pm$ 3.52} & 26.27 \textsuperscript{$\pm$ 4.23} & 57.47 \textsuperscript{$\pm$ 6.62} & 94.98 \textsuperscript{$\pm$ 1.39} & 68.61 \textsuperscript{$\pm$ 1.39} \\
 CharacterBERT   & \textbf{87.98} \textsuperscript{$\pm$ 1.77} & 70.16 \textsuperscript{$\pm$ 3.63} & 69.35 \textsuperscript{$\pm$ 3.01} & \textbf{31.27} \textsuperscript{$\pm$ 9.30} & \textbf{64.19} \textsuperscript{$\pm$ 7.03} & \textbf{96.13} \textsuperscript{$\pm$ 0.70}   & \textbf{69.85} \textsuperscript{$\pm$ 0.70}  \\
\bottomrule
\end{tabular}
\caption{NER average of F1 scores over 5 seeds with the standard deviation with gold tokenization\footnote{We do not report scores for \comp as we only have 3 occurrences in the test set}.}
\label{table:f1-scores-ner-all-tags}
\end{table*}

\paragraph{NER}

Table \ref{table:f1-scores-ner-all-tags} presents the results for NER\footnote{We use Seqeval \cite{seqeval} classification report.}. The CharacterBERT model achieves the highest F1 scores for \loc and \oth categories, as well as the best performance for \perderiv and \perderiva. On the other hand, the DziriBERT model outperforms the other models in the \org and \per categories. It is important to note that the performance varies significantly across the different categories, reflecting the diverse challenges posed by each entity type. For instance, some categories contain named entities with variations of the same word, such as ``Allah''/``Alah''/``Elah'', which translates into God for \perderiva. Since CharacterBERT uses character-level information, it is more robust to noise, which explains the high performances for those entities.

\paragraph{Offensive Language Detection}
The imbalance between non-offensive and offensive instances is challenging during the models' training and evaluation. For example, we fail to train mBERT as it only predicts non-offensive labels corresponding to the majority class. This can also be explained by how hard the distinction between offensive and non-offensive content is without context and external knowledge, as explained before. This also raises the question of how relevant is the backgrounds of the annotators for the offensive detection dataset \cite{basile2020s,uma2021learning,almanea-poesio-2022-armis}.

\begin{table}[!h]
\footnotesize
\centering
\begin{tabular}{lccc}
\toprule
Model & Off & Non-Off& macro avg\\
\midrule
mBERT   & 0.00 \textsuperscript{$\pm$ 0.00} & \textbf{89.73} \textsuperscript{$\pm$ 0.00} & 44.87 \textsuperscript{$\pm$ 0.00}\\
DziriBERT     & \textbf{36.77} \textsuperscript{$\pm$ 10.88} & \textbf{84.78 }\textsuperscript{$\pm$ 2.58} & 60.78 \textsuperscript{$\pm$ 6.21} \\
 CharacterBERT & 24.58 \textsuperscript{$\pm$ 7.44} & 80.21 \textsuperscript{$\pm$ 3.66} & 52.39 \textsuperscript{$\pm$ 3.18}   \\
\bottomrule
\end{tabular}
\caption{Offensive language detection F1 scores, \textit{off} for offensive and \textit{Non-Off} for non offensive}
\label{tab:dataset_offensive_f1}
\end{table}

\section{Discussion}
\subsection{Impact of the Pre-training Corpus}
In Appendix \ref{sec:appendix}, we present the results of all our experiments using the CharacterBERT model trained by \citet{riabi-etal-2021-character}. We observe a heterogeneous improvement in performance, with predominantly better outcomes for our CharacterBERT. We hypothesize that the impact of filtering the training data may not be overly beneficial, possibly due to some smoothing during the training process. Both models' final training data sizes are comparable: 99k for CharacterBERT \cite{riabi-etal-2021-character} and 91k for our CharacterBERT. Nevertheless, we believe this new corpus can be a valuable resource for this language.
\subsection{Impact of Tokenization}
In this section, we investigate the tokenization influence on the enhanced NArabizi Treebank, with a particular emphasis on the homogenization of the tokenization \footnote{We follow the  terminology of UD where a tokenizer performs token segmentation (i.e. source tokens).} and its subsequent impact on our tasks. We also evaluate the models in a realistic scenario where gold tokenization is unavailable. We use the UDPipe tokenizer \cite{straka-etal-2016-udpipe} that employs a Gated Linear Units (GRUs) \cite{cho-etal-2014-learning} artificial neural network for the identification of token and sentence boundaries in plain text. It processes fixed-length segments of Unicode characters and assigns each character to one of three classes: token boundary follows, sentence boundary follows, or no boundary\draftremove{By leveraging GRUs and dropout for regularization, the tokenizer achieves efficient performance with reduced computational requirements}. The tokenizer is trained using the Adam stochastic optimization method, employing randomly shuffled input sentences to ensure effective tokenization across various NLP tasks. 
\begin{table}[h!]
\begin{center}
{\footnotesize
\begin{tabular}{rccc}
\toprule
{ Tokenizer} & { Prec} & { Recall} & { F1} \\
\hline
\rule{0pt}{2.5ex}Tokens & 97.10 \textsuperscript{$\pm$ 0.35} & 95.49 \textsuperscript{$\pm$ 0.45}& 96.29 \textsuperscript{$\pm$ 0.39}\\
Multiwords & 79.74 \textsuperscript{$\pm$ 4.30}& 33.81 \textsuperscript{$\pm$ 2.87}& 47.35 \textsuperscript{$\pm$ 2.59}\\
Words & 92.92\textsuperscript{$\pm$ 0.65} & 88.06 \textsuperscript{$\pm$ 0.96}& 90.42\textsuperscript{$\pm$ 0.80} \\
\bottomrule
\end{tabular}

\caption{Tokenization evaluation average scores over 5 folds }
\label{tab:tokenization_eval}
}
\end{center}
\end{table}

We conduct a 5-fold evaluation using the UDPipe tokenizer and assess its performance based on the token-level, multiword, and word-level scores. The results in Table \ref{tab:tokenization_eval} show high scores for the tokens and words F1 scores demonstrate the tokenizer's efficacy in handling various tokens and words, which shows that the tokenization for NArabizi is learnable. We also notice sub-optimal performance regarding multi-words, due to their random occurrence nature.\footnote{It is important to note that tokens refer to surface tokens (e.g., French ``au" counts as one token), while words represent syntactic words (``au" is split into two words, ``à" and ``le").}.

For our following experiments, we train a tokenizer using the train and dev as held-out and tokenize the test set for evaluation. We do not predict the boundaries of the sentence.

 \begin{table*} [!h]
\centering
\footnotesize
\begin{tabular}{@{}lccc|ccc@{}}
\toprule
\multirow{2}{*} {Model}& \multicolumn{3}{c}{\textit{Predicted tokenization}}                                               & \multicolumn{3}{c@{}}{\textit{NArabiziV1 tokenization}}                                            \\ 
\cline{2-7}
&   \rule{0pt}{2.5ex}{UPOS} & {UAS} & {LAS} & {UPOS} & {UAS} & {LAS}  \\ 
\hline
     \rule{0pt}{2.5ex} mBERT   & 72.44 \textsuperscript{$\pm$ 0.87} & 61.40 \textsuperscript{$\pm$ 0.29} & 50.39 \textsuperscript{$\pm$ 0.64}  & 75.84 \textsuperscript{$\pm$ 0.92} & 65.77 \textsuperscript{$\pm$ 0.40} & 54.15 \textsuperscript{$\pm$ 0.68}  \\
                        \rule{0pt}{2.5ex} DziriBERT  & 76.27 \textsuperscript{$\pm$ 1.46} & 65.35 \textsuperscript{$\pm$ 0.39} & 55.04 \textsuperscript{$\pm$ 0.65}     & 79.49 \textsuperscript{$\pm$ 1.63} & 70.04 \textsuperscript{$\pm$ 0.48} & 59.19 \textsuperscript{$\pm$ 0.70}       \\
                        CharacterBERT   & 70.03 \textsuperscript{$\pm$ 2.10} & 61.08 \textsuperscript{$\pm$ 0.18} & 49.13 \textsuperscript{$\pm$ 0.42}    & 73.10 \textsuperscript{$\pm$ 2.33} & 65.37 \textsuperscript{$\pm$ 0.22} & 52.99 \textsuperscript{$\pm$ 0.50} \\
                         
\bottomrule
\end{tabular}
\caption{UD results for models trained on NArabiziV2 treebank and tested on test set with predicted tokenization and old tokenization from NArabiziV1}
\label{tab:ud_pred_tok}
\end{table*}

\paragraph{Pos-tagging and Dependency Parsing}

Table \ref{tab:ud_pred_tok} presents the results for models trained on the NArabiziV2 training set and tested on both the predicted tokenization and the previous version of tokenization with gold annotations from NArabiziV2. The outcomes for the predicted tokenization indicate that despite having a well-performing tokenizer, as demonstrated in Table \ref{tab:tokenization_eval}, there is still a substantial loss in performance when compared to the gold tokenization results, highlighted by the red box in Table \ref{table:goldtokenparsing}. Similarly, using the tokenization from NArabiziV1 and gold annotations from NArabiziV2 also exhibits a significant drop in performance. This observation first highlights the impact of the corrections brought to standardize the  treebank tokenization and \draftadd{then, given the difference of performance between predicted and gold tokens,  calls for the development of morphological-analysers, crucial for Arabic-based dialects, as  UD tokenization is indeed a morpho-syntactic process.}

\paragraph{Named Entity Recognition Evaluation on Non-Gold Tokenization}

The conventional evaluation methodology for NER typically assigns entities to distinct token positions. Nevertheless, this method proves inadequate when the token count for evaluation differs from the number of gold tokens, which is almost always the case when processing user-generated content. 

As a result, we adopt the evaluation strategy devised by \citet{bareket-tsarfaty-2021-neural}, which associates entities with their forms instead of their indices. This approach yields F1 scores based on strict, exact matches of surface forms for entities, irrespective of the category distinctions, thereby offering a more accurate and reliable evaluation in scenarios with varying token counts. In other words, the gold and predicted NE spans must exhibit an exact match regarding their form, boundaries, and associated entity type.
\begin{table}[!h]
\footnotesize
\centering
\begin{tabular}{lcc}
\toprule
Model & \textit{Gold} & \textit{Predicted} \\
\midrule
mBERT   & 71.79 \textsuperscript{$\pm$ 2.30}  & 66.76 \textsuperscript{$\pm$ 1.52}\\
DziriBERT    & 75.56 \textsuperscript{$\pm$ 2.13}   & 68.89 \textsuperscript{$\pm$ 2.64} \\
 CharacterBERT   & \textbf{76.30} \textsuperscript{$\pm$ 1.29}  & \textbf{70.54} \textsuperscript{$\pm$ 2.00}  \\
\bottomrule
\end{tabular}
\caption{Comparison of NER scores for \per / \loc / \org entities F1 micro average on predicted tokenization and gold tokenization averaged across five seeds.}
\label{tab:dataset_ner_tok}
\end{table}

Table \ref{tab:dataset_ner_tok} presents the NER scores, considering our three main NE categories: \per, \loc, and \org. As expected, we observe a decline in performance when evaluating the models using predicted tokenization. The CharacterBERT model exhibits the best performance on gold and predicted tokenization. Moreover, when evaluated using predicted tokenization, all models demonstrate a similar performance drop. This demonstrates that there is an important gap when evaluating using gold tokenization, which raises the question of how much the current evaluation of NER models reflects the actual model performance in a realistic setting for noisy UGC.

\section{Conclusion}
In this paper, we present a comprehensive study on the development and refinement of the NArabizi Treebank \cite{seddah-etal-2020-building} by improving its annotations, consistency, and tokenization, as well as providing new annotations for NER and offensive language. Our work contributes to the enhancement of the NArabizi Treebank, making it a valuable resource for research on low-resource languages and user-generated content with high variability. We explore the impact of tokenization on the refined NArabizi treebank, employing the UDPipe tokenizer for our evaluation. The results demonstrate the tokenizer's effectiveness in handling various tokens and multiword expressions. Our experiments show that training and testing on the NArabiziv2 improve the UD tasks performances. Furthermore, we show the impact of the tokenization for NER and UD tasks, and we report results using predicted tokenization for evaluation to estimate the models' performance on raw data.     

Future research could emphasize expanding the NArabizi Treebank \draftreplace{to encompass other linguistic phenomena}{towards other dialects} and examining the treebank's potential applications in various NLP tasks. Our dataset is made freely available as part of the new version of the Narabizi Treebank\footnote{\url{https://gitlab.inria.fr/ariabi/release-narabizi-treebank}}.
The next release will additionally contain a set of other sentence translations  prepared by a Tunisian speaker.  These translations will be interesting for cross-dialect studies, given that the Narabizi corpus is predominantly made of Algerian dialect.

\section*{Acknowledgements}
We warmly thank the reviewers for their very valuable feedback.
This work  received funding from the European Union’s Horizon 2020
research and innovation programme under grant agreement No. 101021607. 
We are grateful to Roman Castagné for his valuable feedback and proofreading and wish to gratefully acknowledge the OPAL infrastructure from Université Côte d'Azur for providing resources and support. 

\bibliography{anthology,custom}
\bibliographystyle{acl_natbib}
\appendix

\section{Appendix}
\label{sec:appendix}
\subsection{Datasets}

\begin{table}[!h]
    \centering
    \footnotesize
\begin{tabular}{llll}
\hline
\multicolumn{2}{c}{NERDz}            & \multicolumn{2}{c}{Our dataset}       \\     
\midrule
\multicolumn{1}{l}{Entities} & Count & \multicolumn{1}{l}{Entities} & Count \\     \toprule
\per                           & 467   & \per                           & 479   \\ 
GPE/\loc                       & 479   & \loc                           & 466   \\ 
\org    
& 290   & \org/\comp                      & 265  \\
    \bottomrule
\end{tabular}%
    \caption{Mapping of NER labels in our dataset to the Published NERDz dataset \cite{touileb2022nerdz}.}
    \label{tab:comp-nerdz}
\end{table}
\subsection{Results with CharacterBERT from \cite{riabi-etal-2021-character}}

 \begin{table*} [h!]
\resizebox{.95\textwidth}{!}{%
\centering
\footnotesize

\begin{tabular}{@{}clccc|ccc@{}}
\toprule
\multirow{2}{*} {Model}& \multirow{2}{*}{\backslashbox{\scriptsize{Train}\kern-1.5em}{\kern-1.5em \scriptsize{Test}}} & \multicolumn{3}{c}{\textit{NArabiziV1}}                                               & \multicolumn{3}{c@{}}{\textit{NArabiziV2}}                                            \\ 
\cline{3-8}
&  & \rule{0pt}{2.5ex}{UPOS} & {UAS} & {LAS} & {UPOS} & {UAS} & {LAS}  \\ 
\hline
               \rule{0pt}{2.8ex}   CharacterBERT \scriptsize{\cite{riabi-etal-2021-character}} &\multirow{2}{*}{\rotatebox[origin=c]{90}{\textit{{\centering \scriptsize{NArabiziV1}}}}}  & 75.33 \textsuperscript{$\pm$ 2.77} & 67.86 \textsuperscript{$\pm$ 0.95} & 54.40 \textsuperscript{$\pm$ 0.81} & 72.33 \textsuperscript{$\pm$ 2.60} & 65.17 \textsuperscript{$\pm$ 0.79} & 51.51 \textsuperscript{$\pm$ 1.05} \\
                \rule{0pt}{4.4ex}        CharacterBERT \scriptsize{(Ours)}  &  & \textbf{76.19} \textsuperscript{$\pm$ 2.48} & \textbf{68.78} \textsuperscript{$\pm$ 0.36} & \textbf{55.14} \textsuperscript{$\pm$ 0.38} & \textbf{73.01} \textsuperscript{$\pm$ 2.05} & \textbf{66.10} \textsuperscript{$\pm$ 0.48} & \textbf{52.41} \textsuperscript{$\pm$ 0.50} \\
                  \midrule
  \rule{0pt}{2.8ex} CharacterBERT \scriptsize{\cite{riabi-etal-2021-character}} &\multirow{2}{*}{\rotatebox[origin=c]{90}{\textit{{\centering \scriptsize{NArabiziV2}}}}}   & \textbf{72.46} \textsuperscript{$\pm$ 3.19} & 65.30 \textsuperscript{$\pm$ 0.50} & 51.84 \textsuperscript{$\pm$ 0.68}   & \textbf{79.65} \textsuperscript{$\pm$ 0.90} & \textbf{70.56} \textsuperscript{$\pm$ 0.32} & \textbf{58.08} \textsuperscript{$\pm$ 0.76}    \\
  \rule{0pt}{4.4ex}  CharacterBERT \scriptsize{(Ours)}   &  & 72.24 \textsuperscript{$\pm$ 2.62} & \textbf{65.74} \textsuperscript{$\pm$ 0.24} & \textbf{51.86} \textsuperscript{$\pm$ 0.51}   & 76.34 \textsuperscript{$\pm$ 2.68} & 69.84 \textsuperscript{$\pm$ 0.27} & 56.27 \textsuperscript{$\pm$ 0.54}    \\
\bottomrule
\end{tabular}
}
\caption{Results for UD on test set, DEV set is used for validation (with gold tokenization) (We report average of F1 scores over 5 seeds with the standard deviation)}
\label{table:goldtokenparsing2}
\end{table*}

\begin{table*}[h!]

\resizebox{.95\textwidth}{!}{%
\centering
    \footnotesize
\begin{tabular}{lcccccccc}
\toprule
Model & \loc & \org & \per & \oth & \perderiv & \perderiva \\
\midrule
 \rule{0pt}{2.5ex} CharacterBERT  \scriptsize{\cite{riabi-etal-2021-character}}  & 86.80 \textsuperscript{$\pm$ 2.01} & 68.53 \textsuperscript{$\pm$ 6.09} & 65.36 \textsuperscript{$\pm$ 2.74} & \textbf{45.16} \textsuperscript{$\pm$ 13.60} & 58.96 \textsuperscript{$\pm$ 10.42} & 95.00 \textsuperscript{$\pm$ 1.32}    \\
  \rule{0pt}{2.5ex} CharacterBERT \scriptsize{(Ours)} & \textbf{87.98} \textsuperscript{$\pm$ 1.77} & \textbf{70.16 }\textsuperscript{$\pm$ 3.63} & \textbf{69.35} \textsuperscript{$\pm$ 3.01} & 31.27 \textsuperscript{$\pm$ 9.30} & \textbf{64.19} \textsuperscript{$\pm$ 7.03} & \textbf{96.13} \textsuperscript{$\pm$ 0.70}    \\
\bottomrule
\end{tabular}
}
\caption{NER average of F1 scores over 5 seeds with the standard deviation with gold tokenization\footnote{We do not report scores for \comp as we only have 3 occurrences in the test set}.}
\label{table:f1-scores-ner-all-tags2}
\end{table*}

\begin{table}[!h]
\footnotesize
\centering
\begin{tabular}{lccc}
\toprule
Model & Off & Non-Off& macro avg\\
\midrule
 CharacterBERT  \scriptsize{\cite{riabi-etal-2021-character}} & \textbf{36.29} \textsuperscript{$\pm$ 5.73} & 76.49 \textsuperscript{$\pm$ 3.81} & \textbf{56.39} \textsuperscript{$\pm$ 2.95}   \\
  CharacterBERT \scriptsize{(Ours)} & 24.58 \textsuperscript{$\pm$ 7.44} & \textbf{80.21} \textsuperscript{$\pm$ 3.66} & 52.39 \textsuperscript{$\pm$ 3.18}   \\
\bottomrule
\end{tabular}
\caption{Offensive language detection F1 scores, \textit{off} for offensive and \textit{Non-Off} for non offensive}
\label{tab:dataset_offensive_f12}
\end{table}

\end{document}